\pdfoutput=1

\documentclass[11pt]{article}

\usepackage[final]{acl}

\usepackage{times}
\usepackage{latexsym}

\usepackage[T1]{fontenc}

\usepackage[utf8]{inputenc}

\usepackage{microtype}

\usepackage{inconsolata}

\usepackage{graphicx}
\usepackage{multirow}

\usepackage{enumitem}
\usepackage{xspace}

\newcommand{\myhide}[1]{}

\newcommand{\reminder}[1]{{\textsf{\textcolor{red}{#1}}}}

\newcommand{\todo}[1]{{\textsf{\textcolor{red}{TODO: #1}}}}
\newcommand{\note}[1]{{\textsf{\textcolor{blue}{NOTE: #1}}}}

\newcommand{\mkclean}{
    \renewcommand{\reminder}{\myhide}
    \renewcommand{\notice}{\myhide}
    \renewcommand{\todo}{\myhide}
    \renewcommand{\lmn}{\myhide}
    \renewcommand{\note}{\myhide}
}

\newcommand{\bit}{\begin{compactitem}}
\newcommand{\eit}{\end{compactitem}}
\newcommand{\ben}{\begin{compactenum}}
\newcommand{\een}{\end{compactenum}}

\def\checkmark{\tikz\fill[scale=0.4](0,.35) -- (.25,0) -- (1,.7) -- (.25,.15) -- cycle;}

\newcommand{\ours}{\textsc{Trivia$+$}\xspace}
\newcommand{\trivia}{\textsc{TriviaQA}\xspace}

\newcommand{\llmaaj}{{LLM-as-a-Judge}\xspace}

\newcommand{\dolly}{\textsf{Dolly}\xspace}
\newcommand{\ragtr}{\textsf{RAGTruth}\xspace}
\newcommand{\phd}{\textsf{PHD}\xspace}
\newcommand{\halu}{\textsf{HaluEval}\xspace}
\newcommand{\haluw}{\textsf{HaluEval-Wild}\xspace}
\newcommand{\semeval}{\textsf{SemEval}\xspace}
\newcommand{\facts}{\textsf{FACTS}\xspace}
\newcommand{\summac}{\textsf{SummaC}\xspace}

\newcommand{\hal}{hallucination\xspace}
\newcommand{\hals}{hallucinations\xspace}

\newcommand{\lmn}[1]{{\textsf{\textcolor{teal}{LMN: #1}}}}

\usepackage{booktabs}     

\usepackage{amsfonts}       
\usepackage{nicefrac}       
\usepackage{microtype}      
\usepackage{xcolor}         

\usepackage{tikz}
\usepackage{listings} 
\usepackage[many]{tcolorbox}

\usepackage{ifthen}
\newboolean{shownew}
\setboolean{shownew}{false} 

\newcommand{\newtxthelper}[1]{%
  \ifthenelse{\boolean{shownew}}%
    {\textcolor{blue}{#1}}%
    {#1}%
}
\DeclareRobustCommand{\newtxt}[1]{\newtxthelper{#1}}
\DeclareRobustCommand{\reftag}[1]{}

\lstdefinestyle{custompython}{
  language=Python,
  basicstyle=\ttfamily\small,
  backgroundcolor=\color{gray!10},
  frame=single,
  breaklines=true,
  breakindent=0pt,
  showstringspaces=false,
  keywordstyle=\ttfamily\small,
  commentstyle=\color{gray},
}
\usepackage{footmisc}

\title{Rethinking Evaluation for LLM Hallucination Detection: \\ A Desiderata, A New RAG-based Benchmark, New Insights}

\author{
  \textbf{Wenbo Chen\textsuperscript{1}},
  \textbf{Veena Padmanabhan\textsuperscript{1}},
  \textbf{Tootiya Giyahchi\textsuperscript{1}},
  \textbf{Elaine Wong\textsuperscript{1}},
  \textbf{Leman Akoglu\textsuperscript{1,2}}
\\\\
  \textsuperscript{1}Amazon \quad
  \textsuperscript{2}Carnegie Mellon University \\
  \texttt{\{wbchen, veenapad, tootiya, elawong\}@amazon.com, lakoglu@andrew.cmu.edu}
}

\newcommand{\notice}[1]{#1}
\mkclean
\begin{document}
\maketitle

\begin{abstract}

Hallucination, broadly referring to unfaithful, fabricated, or inconsistent content generated by LLMs, has wide-ranging implications. Therefore, a large body of effort has been devoted to detecting LLM hallucinations, as well as designing benchmark datasets for evaluating these detectors.
In this work, we first establish a \textbf{desiderata} of properties for \hal detection benchmarks (HDBs) to exhibit for effective evaluation.
A critical look at existing HDBs through the lens of our desiderata reveals that \textit{none of them exhibits all the properties}. We identify two largest gaps: \textbf{(1)}  \textbf{RAG-based} grounded benchmarks with {long context} are severely lacking (partly because length impedes human annotation); and \textbf{(2)} Existing benchmarks do not make available realistic \textbf{label noise} for stress-testing detectors although real-world use-cases often grapple with {label noise}  due to human or automated/weak annotation. 
To close these gaps, we build and open-source \textbf{a new RAG-based HDB} called \ours that underwent a rigorous human annotation process. Notably, our benchmark exhibits all desirable properties including \textbf{(1)} \ours contains samples with the longest context in the literature;  and \textbf{(2)} we design and share \newtxt{four} sets of noisy labels with different\newtxt{, both sample-dependent and sample-independent,} noise schemes.\reftag{KaS4-P5/Zsjz}
Finally, we perform \textbf{experiments on RAG-based HDBs}, including our \ours, using popular SOTA detectors that  reveal \textbf{new insights}: \textbf{(i)} 
 ample room remains for current detectors to reach the  performance ceiling on RAG-based HDBs, \textbf{(ii)} 
the basic \llmaaj baseline performs competitively, and  
 \textbf{(iii)} label noise hinders detection performance.
We expect that our findings, along with our proposed benchmark\footnote{We release the benchmark dataset with expert annotations at \href{https://github.com/amazon-science/hallucination-benchmark-trivialplus}{github.com/amazon-science/hallucination-benchmark-trivialplus}.}, will motivate and foster needed research on hallucination detection for RAG-based tasks. 
\end{abstract}

\section{Introduction}
\label{sec:intro}





Modern large language models (LLMs) have propelled advances in various real-world domains including  e-commerce \cite{jiang2024hallucination},  medicine \cite{thirunavukarasu2023large}, law \cite{thomson,LexisNexis}, to name a few.
While LLM-driven AI technologies offer
real-world impact, 
the fact that they \textit{hallucinate} remains
a big obstacle in the safe and secure  usage of generative AI tools \cite{applesuspends,magesh2024hallucination,hong2024hallucinations}.


Various efforts have been devoted to preventing LLM \hals.
Mitigation strategies include  retrieval-augmented generation (RAG) \cite{li2024enhancing}, reasoning \cite{conf/acl/DhuliawalaKXRLC24}, self-reflection \cite{ji2023towards}, self-refinement \cite{madaan2023self}, among others \cite{zhang2023siren}. 
Nevertheless, \hals persist 
making both proactive mitigation and post hoc detection essential \cite{luo2024hallucination}. As a result, the field has
seen a surge of interest on \hal detection approaches \cite{huang2025survey,luo2024hallucination,zhang2023siren}
and a large body of \hal detection benchmark (HDB) datasets have also been developed (see Table \ref{tab:compare}).

Despite the long list of HDBs, we point to a large gap: \textbf{only a handful of RAG-based HDBs exists in the literature}. These HDBs are especially hard for humans to annotate, as they  (1) exhibit considerably \textit{long context} settings, and (2) associate with \textit{knowledge-intensive} tasks \cite{lewis2020retrieval}. 
In this work, we build and open-source
a new RAG-based HDB called \ours; representative of knowledge-intensive LLM tasks and thus grounded on often long context---rendering \hal detection even more challenging. Furthermore, we identify characteristics that a HDB ideally exhibits. 
While admittedly not exhaustive, the list offers a systematic lens to assess existing HDBs. 
Lastly, experiments on RAG-based HDBs report new findings. The following summarizes the contributions of this work.

\vspace{-0.1in}
\begin{itemize}
[align=left, leftmargin=1.2em, 
labelindent=3pt,listparindent=0pt, labelwidth=0.5em, itemindent=!,itemsep=-3pt]

    \item \textbf{Desiderata for HDBs (Hallucination Detection Benchmarks):~} We compose a list of desirable properties for a HDB  to exhibit, through the lens of which 
    we scrutinize the existing HDBs---revealing that none of them satisfies all the criteria in our desiderata. (See Table \ref{tab:compare}.)
    
    \item \textbf{A New Benchmark:~} We introduce \ours, a new RAG-based HDB that exhibits seven key properties in our desiderata. Namely, it consists of \textbf{(i)}  \textit{organic}, naturally-occurring \hals; \textbf{(ii)}  \textit{human-verified} labels, with each sample annotated up to 6 times;
    \textbf{(iii)}
    \textit{long-context} (i.e. high-demand, hard-to-label) LLM tasks;
 \textbf{(iv)} 
       training \textit{labels with realistic noise} (for supervised detectors); 
\textbf{(v)} \textit{extrinsic and intrinsic} \hals; from
 \textbf{(vi)}  \textit{multiple} LLMs; and
 \textbf{(vii)}  \textit{multiple} domains. 

    \item \textbf{Empirical Findings:} Experiments on RAG-based HDBs  find that \textbf{(1)} hallucination prevails in grounded tasks and current detectors leave ample room to ceiling performance,  \textbf{(2)} while supervised and fine-tuned models can be effective, the simple
\llmaaj performs competitively, in contrast to reports in past literature \cite{conf/acl/NiuWZXSZS024},\reftag{KaS4-S7}
and \textbf{(3)} sample-dependent (as opposed to random) label noise   hinders detection significantly. 

    \item \textbf{Open Problems:~} Our findings highlight the scarcity of \textbf{(a)} HDBs with long context---and effective detectors on RAG-based generations, 
    \textbf{(b)} HDBs with  realistic label noise---and detectors leveraging robust learning; and \textbf{(c)} HDBs that span from multiple domains---and detectors that can generalize across domains.
\end{itemize}

\section{A Desiderata for Hallucination Detection Benchmarking}
\label{sec:benchmark}
\begin{table*}[!t]
\centering
\caption{Comparison of HDBs w.r.t. desirable desiderata (Sec. \ref{sec:benchmark}). Proposed \ours exhibits all seven properties; with \textbf{(i)} organic generations, \textbf{(ii)} human-verified labels (for evaluation), \textbf{(iii)} \newtxt{four} sets of realistic label noise (for training),\reftag{KaS4-P5/Zsjz}
on \textbf{(iv)} RAG-based long-context tasks (knowledge-intensive, hard-to-annotate), with \textbf{(v)} both extrinsic and intrinsic \hals, from \textbf{(vi)} multiple modern LLMs and \textbf{(vii)} multiple domains. \textbf{None of the existing HDBs fully meets the desiderata.} RAG-based HDBs are few, and are highlighted in \textcolor{red}{red}. Those marked with $^\times$ do not make LLM generations available. Symbol ?  denotes that it is unclear whether the dataset spans multiple domains. 
}
\vspace{-0.1in}
\scalebox{0.7}{
\begin{tabular}{ll|cc|cc|ccc}
\toprule
&                                
& \textbf{Hal.s:} 
& \textbf{Test-Labels:} 
& \textbf{Train-Labels:} 
& \textbf{Context:} 
& \textbf{Type:} 
& \textbf{LLMs:}  
& \textbf{Domains:} \\
\textbf{\#} 
& \textbf{Dataset}    & \textbf{Organic} & \textbf{Human} & \textbf{Realistic noise} & \textbf{Long/RAG} & \textbf{Faith.} & \textbf{Multiple}  & \textbf{Multiple} \\
\midrule \midrule
1   & HADES \cite{conf/acl/liu2021token}     &   & \checkmark  &  &  &  &  & ? \\  \hline
2   & Frank \cite{conf/acl/marfurt-henderson-2022-unsupervised}   &   \checkmark     &                        &  &  & \checkmark &  &  \\  \hline
3   & TLHD-CNNDM  \cite{conf/acl/marfurt-henderson-2022-unsupervised}    &   \checkmark     & \checkmark                       &  &  & \checkmark &  &  \\  \hline
4   & SummaC \cite{conf/tacl/laban-etal-2022-summac} & \checkmark     & \checkmark                      &  &  &  \checkmark&  & \checkmark \\  \hline
5   & WikiBio+ \cite{conf/emnlp/ManakulLG23} & \checkmark      & \checkmark                      &  &  &  &  &  \\ \hline
6   & \textcolor{red}{HaluEval} \cite{conf/emnlp/li-etal-2023-halueval} &  &                  &  & \checkmark & \checkmark &  & \checkmark \\  \hline
7  & HILT \cite{conf/emnlp/rawte-etal-2023-troubling} & \checkmark$^\times$     & \checkmark                     &  &  &  & \checkmark  & ?  \\  \hline
8  & PHD \cite{conf/emnlp/YangS023} & \checkmark     & \checkmark   &  &  &  &  & \checkmark  \\ \hline
9  & \textcolor{red}{DelucionQA} \cite{sadat2023delucionqa} & \checkmark$^\times$     &\checkmark &  & \checkmark & \checkmark &  &  \\  \hline
10 & \textcolor{red}{RAGTruth} \cite{conf/acl/NiuWZXSZS024} & \checkmark      & \checkmark                       &  & \checkmark & \checkmark &  \checkmark & \checkmark \\  \hline
11 & TofuEval \cite{conf/naacl/tang-etal-2024-tofueval} & \checkmark      &\checkmark                      &  &  & \checkmark & \checkmark & ? \\  \hline
12 & FAVABench \cite{conf/colm/mishra2024finegrained} & \checkmark       & \checkmark                      &  &  &  & \checkmark & \checkmark \\  \hline
13  & SHROOM \cite{conf/semeval/MickusZVVTSRA24} 
    & \checkmark    & \checkmark                      & $\ast$ &  & \checkmark & \checkmark  & \checkmark \\ \hline 
14  & DiaHalu \cite{conf/emnlp/chen-etal-2024-diahalu} & \checkmark     & \checkmark                      &  &  & \checkmark & \checkmark & \checkmark \\  \hline
15  & ERBench \cite{conf/neurips/oh2024erbench} & \checkmark    & auto &  &  &  & \checkmark & \checkmark \\  \hline
16  & Dolly (AC) \cite{conf/emnlp/hu-etal-2024-knowledge} & \checkmark    & \checkmark                      &  &  & \checkmark & \checkmark & \checkmark \\  \hline
17  & \textcolor{red}{Dolly (NC)}  \cite{conf/emnlp/hu-etal-2024-knowledge} & \checkmark     & \checkmark                     &  &  \checkmark & \checkmark & \checkmark &  \checkmark\\  \hline
18 & ANAH \cite{ji-etal-2024-anah} & \checkmark      & \checkmark &  &  & \checkmark &  \checkmark & \checkmark \\  \hline
19  & HaluEval-Wild \cite{zhu2024halueval} & \checkmark$^\times$    &                        &  &  &  &  & \checkmark \\  \hline
20  & \textcolor{red}{FACTS} \cite{jacovi2025facts} & \checkmark$^\times$    &                       &  & \checkmark & \checkmark & \checkmark & \checkmark \\  \hline
21  & Mu-SHROOM \cite{vazquez2025semeval} 
& \checkmark    & \checkmark                      & $\ast$ &  &  &  \checkmark& \checkmark \\ \hline
22 & FaithEval \cite{ming2025faitheval} &    & \checkmark & &  & \checkmark &  \checkmark& \checkmark \\ \hline
23 & FaithBench \cite{bao-etal-2025-faithbench} & \checkmark & \checkmark & &  & \checkmark &  \checkmark& \checkmark \\

\midrule
\midrule
24 & \textbf{\textcolor{red}{\ours}} (this paper)  & \checkmark & \checkmark & \checkmark & \checkmark & \checkmark & \checkmark & \checkmark \\
\bottomrule
\end{tabular}
}
\label{tab:compare}
\end{table*}


We start with the question: \textit{What properties should a HDB exhibit?}
We identify  seven {desirable} properties that we discuss and motivate in this section.
While our list may not be exhaustive, it provides a solid foundation on which we scrutinize existing HDBs in the next section.

\begin{figure*}
    \centering
    \begin{tabular}{c|cc}
         \includegraphics[width=0.3\textwidth]{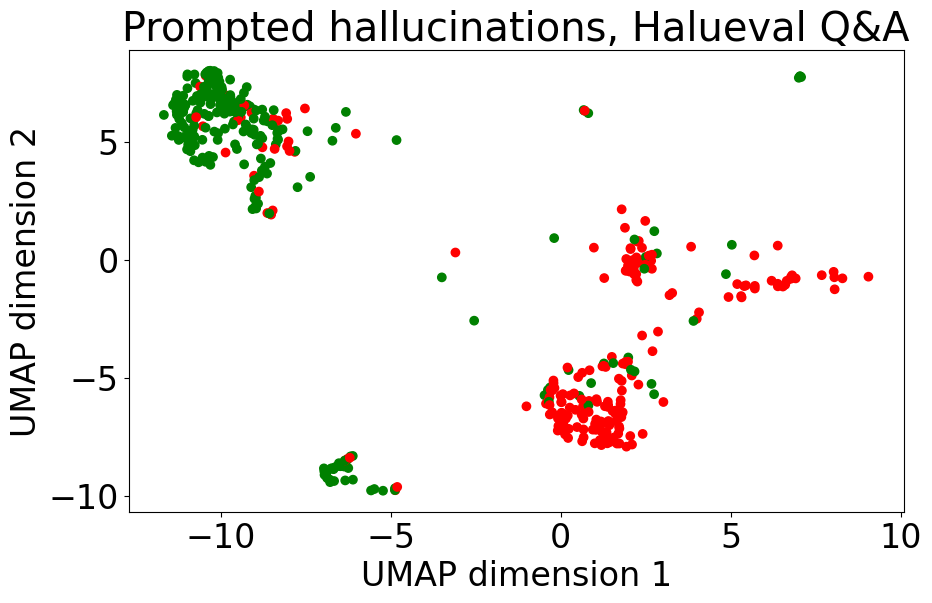}&
\includegraphics[width=0.3\textwidth]{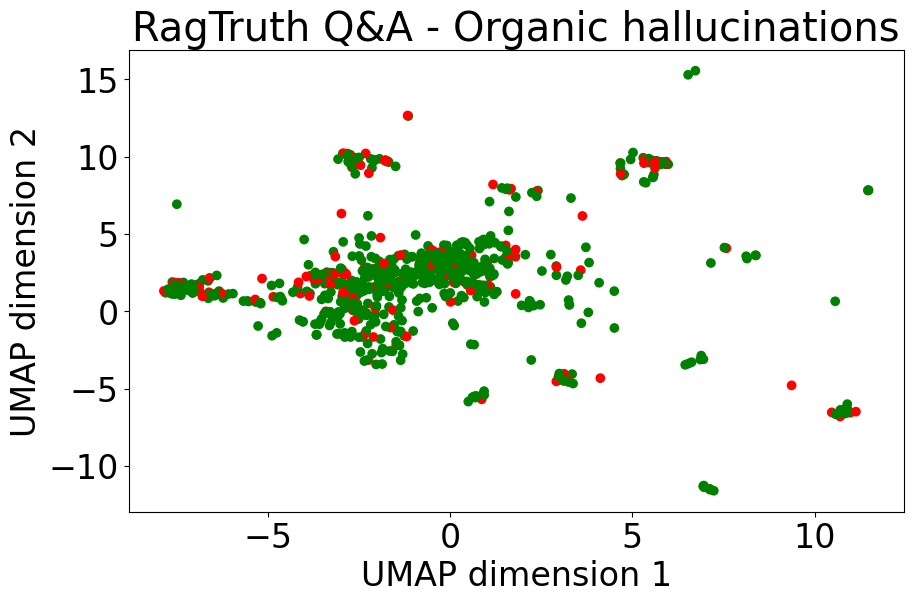} &
\includegraphics[width=0.3\textwidth]{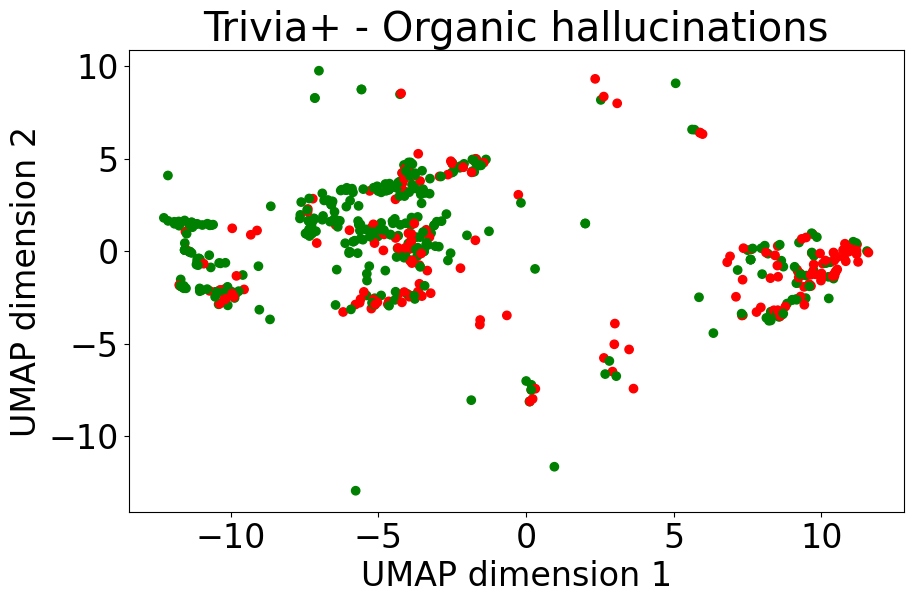}  \\    
    \end{tabular}
        \caption{(best in color) \sloppy{Supervised test split UMAP embeddings generated by fitting on the train split using \texttt{MoritzLaurer/DeBERTa-v3-large-mnli-fever-anli-ling-wanli} \cite{laurer2022less}.  }\textbf{(left)} \textbf{prompted} positive samples (i.e. \hals) in red in \halu stand out away from negatives (green); whereas \textbf{organic} \hals in \ragtr \textbf{(center)}  and \ours \textbf{(right)}  blend in with and resemble the negatives. \newtxt{This visual separability is supported quantitatively in Table~\ref{tab:detect}: SFT achieves 99.6\% F1 on non-organic \halu but only 66--69\% F1 on organic benchmarks.}\reftag{JVop-W2/S2}
         }
         \label{fig:organic}
\end{figure*}

\vspace{-0.1in}
\begin{enumerate}[align=left, leftmargin=0.2em, 
labelindent=3pt,listparindent=0pt, labelwidth=0.5em, itemindent=!,itemsep=-1pt]

\item[\textbf{D1.}] \textbf{Organic (i.e. real, naturally-occurring) Hallucinations:~}
An HDB should ideally contain content organically generated by an LLM, i.e. on its \textbf{natural responses}.
In contrast, hallucinations injected directly (manual) or indirectly (via prompting an LLM to hallucinate) are considered \textbf{non}-organic. In general, we find that despite looking similar to a human eye, non-organic hallucinations are easier to detect and serve as deceptive benchmarks (see Figure  \ref{fig:organic}).

\item[\textbf{D2.}] \textbf{Human-Verified, Reliable Test Labels:~}  To keep a fair and accurate record of progress on a task, it is paramount for benchmark datasets to be equipped with trustworthy labels for evaluation. 
In the ideal scenario, labels provided by \textit{many}, \textit{expert} \textit{human} annotators with \textit{perfect agreement} would be considered \textbf{gold-grade} ground-truth. Label collection is particularly challenging due to (i) the presence of \textit{long-context} samples (e.g. RAG) (ii) the \textit{knowledge-intensive} task of reviewing and absorbing complex-content and (iii) the presence of \textit{subtle} hallucinations (especially in long-form responses) akin to a ``needle'' in the ``haystack'' (i.e. long context),  possibly eluding even the most vigilant annotators. Nevertheless, due diligence should be undertaken to reach as trustworthy and gold-grade labels as possible---with full transparency on the process, regarding the count, expertise, 
and dis/agreement of the annotators.
\newtxt{We note that while human-verified labels are the gold standard, they are costly and time-consuming to obtain, especially for long-context RAG tasks. LLM-as-a-judge labels (e.g. \cite{ji-etal-2024-anah}) offer a practical but noisy alternative, further motivating D4.}\reftag{KaS4-P1}

\item[\textbf{D3.}] \textbf{Long Context (high-demand yet hard-to-label) Tasks:~}
Among the many tasks LLMs are employed on, a HDB should consider tasks ($i$) that are practically-useful and \textit{popular} (i.e. in high demand), 
while ($ii$) potential \hals are \textit{hard} for humans (even experts) to quickly identify.
Then, such a benchmark would contribute to keeping a healthy record of most effective detectors on such high-demand, high-value tasks.
RAG-based long context tasks fall exactly under this category with little emphasis in the HDB literature.

\item[\textbf{D4.}] \textbf{Realistic Training Labels:~} 
While several \hal detectors are unsupervised due to the laborious label gathering involved, some utilize labels. Labels are obtained through various means, including ($i$) employing an \llmaaj (or other unsupervised detectors) or ($ii$) manual labeling, i.e. human annotation.
Depending on the count and expertise of the labelers, one may end up with \textbf{silver-grade} labels. 
In either case, the labels would be \textbf{noisy}.
Further, the labels would exhibit systematic, \textit{sample-dependent} noise, as opposed to random noise. Then, it would be beneficial for an HDB to alleviate the label-collection burden on detection teams by incorporating labeled training data for downstream detectors, while allowing the training data to exhibit \textit{varying degrees of label quality} in line with the aforementioned realistic settings.
While (semi-/supervised) detection teams may split the gold-grade or  ``clean'' evaluation data into train/test for their purposes, we remark that a \textit{noisy} training data would better reflect reality.

    \item[\textbf{D5.}] \textbf{Comprehensive Hallucination Types:} 
    LLM \hals are defined in various ways across the literature, broadly referring to unfaithful, fabricated, inconsistent, or irrational content. We adopt the terminology: \textbf{intrinsic} and \textbf{extrinsic} hallucinations. Former refers to outputs that are \textbf{inconsistent} with the provided reference context.
Latter refer to outputs that are \textbf{unverifiable} by the context.
Others have termed both types in our definition as \textbf{faithfulness} hallucination,  while used the term \textbf{factuality} hallucination
for content that is factually incorrect or fabricated \cite{huang2025survey}.
In comparison, we categorize factual errors that contradict the source context as intrinsic, while those with no basis in external knowledge  as 
extrinsic hallucination since by induction they also have no basis in the reference context. The differences in terminology  aside, a HDB ideally represents different types of hallucinations. 

\vspace{-0.1in}
\item[\textbf{D6.}] \textbf{LLM Diversity:~} It is important to derive organic, natural responses from a number of different LLMs, 
to keep an unbiased record of progress in solving the detection task.
Furthermore,  a benchmark is as relevant as the LLMs it has employed, therefore it is desirable to consider modern and/or popular LLMs.
Admittedly, this is a moving target/goal given the fast-pace of evolution today's LLMs go through, but a relevant one nevertheless.

    \item[\textbf{D7.}] \textbf{Multiple Domains:~} Hallucination detection is relevant in many scenarios as LLMs are employed in diverse domains. As such, detectors that can generalize across domains become critical.  HDBs that provide multi-domain generations stress-test and promote domain generalization.

\end{enumerate}

\section{Existing Benchmarks -- A Critical Look}
\label{sec:data}

Our desiderata provides us with an analytical lens through which we can take a critical look at the existing \hal detection benchmarks and highlight their strengths and weaknesses.
This section elaborates on desirable benchmark properties, as organized into three: Core Properties (\textbf{D1}--\textbf{D2}), Largest Gaps in literature (\textbf{D3}--\textbf{D4}), and Diversity Considerations (\textbf{D5}--\textbf{D7}). 
Table \ref{tab:compare} provides a summary, illustrating the gaps in the literature.

\subsection{Core Properties: \textbf{D1.} Organic Generations \& \textbf{D2.} Verified Test Labels} 
\label{ssec:core}

The core properties in our desiderata are: \textbf{Organic} (text) and \textbf{Human-verified} (labels). These are arguably the two \textit{must-have} criteria that any \hal benchmark should exhibit. Organic content represent naturally-generated (vs. contrived) \hals, while human-verified labels offer trustworthy evaluation results and a fair record of progress on \hal detection.

Table \ref{tab:compare} shows that not all benchmarks readily satisfy these core criteria. 
\textit{Why do we have  benchmarks with non-organic \hals?} 
When recruiting human-annotators is prohibitive, the typical practice is to go backwards and inject \textit{known} hallucinations: (1) directly; by manual perturbation\footnote{Given LLM-generated text, few manually-selected words are masked and re-sampled (varying temperature) from the LLM to yield possibly hallucinated output. As such, 
generation is intervened rather than natural.} \cite{conf/acl/liu2021token} or (2) indirectly; by dictating (i.e. prompting) an LLM to hallucinate \cite{conf/emnlp/li-etal-2023-halueval}.
However,  known injected \hals  offer only the \textit{illusion} of control on label quality, since LLMs may fail to follow instruction when prompted to create \hals---yielding false-positive label noise, while the ``negative'' samples may contain (organic) \hals---yielding false-negative label noise. Further, prompted LLM \hals differ  from organically-generated ones, as 
Figure \ref{fig:organic} illustrates, 
and thus may have limited representation of \textit{natural} \hals. 



\myhide{
\begin{enumerate}[align=left, leftmargin=1.2em, 
labelindent=5pt,listparindent=0pt, labelwidth=0.5em, itemindent=!]

\item  \textbf{[~Non-organic, (Semi-)verified~]} \textit{(LLM-)text-\textbf{perturbed}-to-inject-\hal}: 
Given LLM-generated content, a few manually-selected words are masked and resampled (with varying temperature) from the LLM to yield a different, hallucinated output. As such, while the LLM is involved in generating the \hals, the process is not natural but intervened.
Regarding labels, the ``negative'' samples may be organic and may in fact contain (organic) \hals; i.e. 
the data may exhibit false-negative label noise and should rather be treated positive-unlabeled. 


    \item \textbf{[~Non-organic, Non-verified~]} \textit{LLM-\textbf{prompted}-(not)-to-hallucinate}:
    The given LLM-generated content is fed into an LLM along with a prompt that instructs it to inject hallucinations into it. As such, while the possible \hals are LLM generated, the process is not natural but ``dictated''. Regarding labels,  
    prompt-generated samples may not always contain \hals (as LLM may fail to follow instruction), thus the data possibly exhibits false-positive {label noise}.
   Moreover, similar to the above,  the ``negative'' samples may in fact contain (organic) \hals; i.e. 
the data may also exhibit false-negative label noise.   
     \note{(!): we cannot use these in experiments since no clean-test-labels to properly evaluate? or we do best-effort evaluation and earmark/footnote the results with a grain-of-salt?}
\end{enumerate}

Other benchmarks have adopted organic LLM text, and employed weak (e.g. unsupervised) labelers such as \llmaaj. However, these benchmarks are rarely useful since they cannot be used to fairly and conclusively evaluate detector performance.

\begin{enumerate}[resume,align=left, leftmargin=1.2em, 
labelindent=5pt,listparindent=0pt, labelwidth=0.5em, itemindent=!]
     \item \textbf{[~Organic, Non-verified~]} \textit{\textbf{LLM-labeled}-post-hoc}: These benchmarks contain non-intervened, naturally occurring LLM outputs, but contain labels derived from potentially the same or another, stronger (e.g. proprietary and/or billions-scale) LLM for evaluation. As the LLM labels are not human-verified, and \lmn{since \llmaaj is not as strong a \hal detector as humans}\todo{citation?}, they exhibit label noise (to an unknown degree). 
\end{enumerate}
}


\subsection{Largest Gaps:  \textbf{D3.} RAG-based Tasks \& \textbf{D4.} Realistic Train Labels} 
\label{ssec:gaps}

While diligently human-verified labels are essential for valid evaluation, they are costly to obtain. As a result, various  supervised approaches in the literature often resort to data augmentation with pseudo or proxy labels for training, including several top teams on the SemEval 2024 SHROOM challenge\footnote{\url{https://helsinki-nlp.github.io/shroom/2024}} \cite{chen2024opdai,rykov2024smurfcat,borra2024malto}.
Most common practice is to employ an LLM as a ``judge'', sometimes cross-checking consistency between multiple LLMs \cite{journals/corr/abs-2405-13684,jacovi2025facts}.\reftag{KaS4-S6} However, the degree of noise these training datasets exhibit are often unknown and the vulnerability of detectors to noisy training labels has not been systematically studied.

Alarmingly, Zhu {\em et al.}  show that  modern models
like BERT are (1) quite vulnerable to \textit{sample-dependent} label noise from weak supervision, despite being robust against \textit{random} noise; and that (2) noise-handling methods do not always
improve its performance and may even deteriorate it \cite{ZhuHZAK22}.  Their study focused on text classification, while there exists no study on susceptibility of \hal detectors to label noise from weak supervision. Further, existing work on robust text classification with language models do not consider \hal detection \cite{conf/icnlsp/AgroA23,conf/acl/QiTQXQ23,conf/emnlp/ChongHM22}.

These motivate the necessity of benchmarks with realistic, \textit{sample-dependent} noisy training labels. The current literature, however, falls short in offering a common ground to stress-test existing detectors and thereby fostering research on LNL (learning with noisy labels) in this problem context.
In fact, this is the largest gap we find in the literature as Table \ref{tab:compare} highlights.\footnote{In the table, 
$\ast$ depicts that  SemEval benchmarks do not originally provide noisy labels, yet, individual labels from 5 annotators are shared, which can be used to apply noising strategies \cite{conf/emnlp/ChongHM22} to their final training labels.}

The second major gap is the shortage of available benchmarks that involve RAG-based LLM generations, 
which limits their ability to reflect how modern LLMs are typically used---often heavily augmented with
 RAG for accessing relevant, up-to-date information to create 
accurate, contextually rich, and grounded responses.
The few that are RAG based have short context, that is up to 3$\times$  smaller than \ours on average (see Table \ref{tab:stats})---falling short in representing the advances in efficient, long-context LLMs. 

\subsection{Diversity: \textbf{D5.} Hallucination Types, $\;$\textbf{D6.} LLMs, $\;$\textbf{D7.} Domains} 
\label{ssec:diversity}

Ideally a benchmark should be representative of  hallucinations of various nature, various modern LLMs in popularity, as well as diverse domains they are used in.
Most earlier ``closed-book'' benchmarks focus on factuality of LLM responses\newtxt{, with approaches such as FActScore \cite{min2023factscore} evaluating factual precision at a fine-grained, atomic claim level}.\reftag{KaS4-U3} On the other hand, faithfulness becomes relevant for context-driven tasks, such as summarization, paraphrasing, RAG-based QA. Unfaithful content can be inconsistent with (intrinsic) or unverifiable from (extrinsic) the given context, which \textit{subsumes} factual errors in QA tasks assuming accurate context---thus involving various types of \hals. 

We see in Table \ref{tab:compare} that while the majority of existing benchmarks consider faithfulness, include multiple LLM generations\footnote{In the table, \checkmark$^\times$ depicts HDBs  which do not make available the LLM generations as well as the  labels, including  FACTS,  HaluEval-Wild and DelucionQA.\label{note1}} from various domains, fewer than half satisfy all 3 criteria simultaneously.

\myhide{
\lmn{IDEA: look at openreviews of these papers, if any, to deduce shortcomings}
\todo{Rather than scrutinize existing benchmarks 1-by-1, let's cluster them based on shared-properties in Table 1. (we can keep the enumerated list below for our own eyes/reference.)}

\todo{add links to code repo, if exists.}

\begin{enumerate}

\item  \phd \cite{conf/emnlp/YangS023}, generated by ChatGPT and annotated by human annotators

\item \halu \cite{li2023halueval}\footnote{\url{https://github.com/RUCAIBox/HaluEval}}

\item \haluw
\cite{zhu2024halueval} [“Prompted”] conventional NLP QA/summarization/etc. vs. real-world user-LLM interaction datasets (task-driven vs. dialog)
    \item  \ragtr \cite{conf/acl/NiuWZXSZS024} [organic hallucinations, human labeled]
    
    \item \semeval \cite{conf/semeval/MickusZVVTSRA24}, 2024\footnote{\url{https://helsinki-nlp.github.io/shroom/2024.html}} and 2025\footnote{\url{https://helsinki-nlp.github.io/shroom/}}

\item \facts \cite{jacovi2025facts} Google’s new benchmark [organic hallucinations, LLM labelled, low F1 score]

\item \summac \cite{} summary inconsistency detection

 \item from AlignScore: ``6 popular factual consistency evaluation
datasets, namely XSumFaith, SummEval, QAGS-XSum, QAGS-CNNDM, FRANK and SamSum  

\end{enumerate}

 \note{Detection papers (list below) are/do NOT propose benchmarks (?), but we can see on which datasets they are evaluated, to find more benchmarks.}
  
\begin{enumerate}
   \item Azaria Mitchell [“Prompted“ true/false datasets created by asking an llm to create it]
    \item Universal truth directions [public datasets that aren’t hallucinations e.g. TruthFulQA, incentivized]
\end{enumerate}
}

\section{Proposed Benchmark: \ours}
\label{sec:proposed}

We introduce \ours, a novel dataset contributing to the need for RAG-based HDBs with natural \hals and human-validated labels. We provide a detailed description of the contents and the label annotation process as follows.
\begin{table*}[!th]
\centering
\caption{Stats for RAG-based HDBs. \ours exhibits the longest context samples with higher domain diversity.
}
{\scalebox{0.7}{
\begin{tabular}{llllllcc}
\toprule
  & \multicolumn{3}{c}{\centering{\textbf{context length (\#char.s)}}} & \#\textbf{samples} & \%\textbf{hal.s}& \#\textbf{LLMs} & \textbf{domains}  \\
   \cmidrule(lr){2-4} 
   Benchmark            & median      & mean      & max     & &    &  &           \\\midrule
 \halu (QA)             &     321        &   344        &    1557    &  20K & 50 &   1      &      Multi-hop QA (\texttt{HotPotQA}-based)    \\
 \ragtr (QA)        &   1.2K          &      1.3K     &    2.8K     &   989   & 29.1 &   6   &      Web searches (\texttt{MS-MARCO}-based)     \\
 \dolly (NC)      &  2.97K    &   3.1K          &      5.99K    & 100      & 44.5 & 7 &          Web searches (\texttt{MS-MARCO}-based)          \\\midrule
\ours        &   2.8K   &   9.3K          &  94K         &  3224   & 35 &  3 &          Paragraph reasoning, Web Searches, Medical docs, Wikipedia      \\                              \bottomrule
\end{tabular}
}}
\label{tab:stats}
\end{table*}


\subsection{Description}
\ours contains generations from three less-represented but popular LLMs: A commercially available SOTA LLM, Gemma-7b, and Mixtral 8x7b. For domain diversity, we source prompts from multiple established datasets, specifically \trivia \cite{conf/acl/JoshiCWZ17}, NaturalQuestions \cite{2019nq}, MS-MARCO\cite{DBLP:journals/corr/NguyenRSGTMD16}, CovidQA\cite{moller2020covid} and DROP\cite{Dua2019DROP}. Each prompt contains reference material, pulled directly from the source dataset, and a question (or query). Our dataset construction methodology employs a strategic filtering approach: We first query the commercially available SOTA LLM\newtxt{\footnote{\newtxt{Legal constraints prevent us from naming the specific LLM other than few tech-specs; $>150$B parameters, 2024 release. Labeling methodology is model-agnostic and findings should generalize.}}},\reftag{KaS4-S1}
the largest of the three, with questions from these datasets and use ROUGE Score \cite{lin-2004-rouge} comparisons between ground truth answers and generated responses to identify low-similarity cases (with a filter for similarity scores < 0.1). \newtxt{Our hypothesis is that low ROUGE overlap between the generated answer and the ground-truth answer is likely to correlate with a higher hallucination rate, as low-overlap answers are more likely to deviate from the expected content. This filtering serves as a resource-efficient strategy to enrich the proportion of true hallucinations given limited annotator resources, without altering LLM outputs—thereby preserving their organic nature.}\reftag{KaS4-P2, C1} Using the same context-question pairs, we then prompt the two other LLMs, creating annotation triplets comprising context, question, and answers from all three models. Each triplet receives multiple labels from annotators to ensure label reliability, with specifics of the annotation process described below.

\subsection{Human Annotation}

\begin{figure}
\centering
\includegraphics[width=0.4\textwidth]{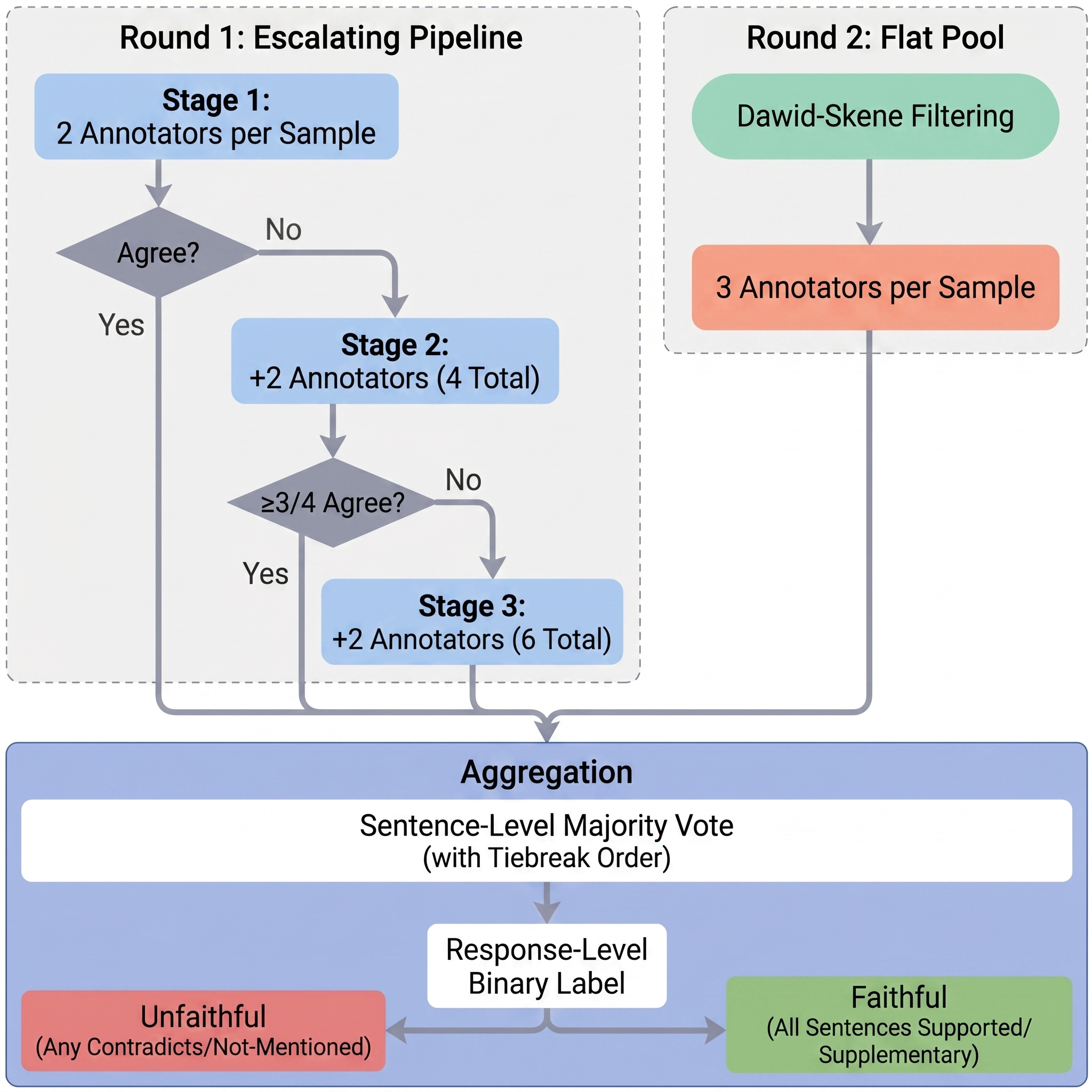}
\caption{\newtxt{Overview of the annotation pipeline. Round~1 uses an escalating strategy: each sentence starts with 2 annotators and escalates to 4 or 6 upon disagreement. Round~2 employs a flat pool of 3 annotators per sample after filtering low-quality workers via the Dawid-Skene model. Sentence-level labels are aggregated by majority vote, then mapped to a binary response-level label.}}
\label{fig:annotation_flow}
\end{figure}

\textbf{Labels \& Definitions:~}
We instructed annotators to provide labels at a sentence level, with each sentence receiving one of four labels: \newtxt{`Supported', `Contradicted', `Not Mentioned' and `Supplementary'}, closely aligning with the definition used by FACTS \cite{jacovi2025facts} . The definition aligns with faithfulness\newtxt{:} \newtxt{`Not Mentioned'} and \newtxt{`Contradicted'} labels correspond to `unfaithful'\newtxt{,} and \newtxt{`Supported'} and \newtxt{`Supplementary'} correspond to `faithful'.\reftag{C6} More details are included in Apdx. \ref{appx:human_guidelines}.

\noindent
\textbf{Annotators:~}
Two rounds of training were performed with a pool of 18 of annotators fluent in English. The team's annotations were audited by the authors to provide detailed feedback in cases where systematic error was observed. At the end of two rounds of training, we measured average accuracy when two annotators agreed on an annotation at 89\% but individual accuracy was about 77\%. We conjecture that our dataset was particularly challenging for human annotators due to long-context examples. We gathered multiple votes per sample from annotators to boost accuracy through aggregation (described below).

\subsubsection{Multi-Vote Annotation for Evaluation Test Labels}
\label{sssec:clean}
As illustrated in Figure~\ref{fig:annotation_flow}, the annotation task was divided into two rounds and was performed at the sentence level. In the first round, we utilized a multi-stage pipeline. Two annotators labeled each sample. When disagreement was observed, two additional annotators provided labels. If there still was no clear majority in voting (three out of four labels being consistent), two additional labels were gathered. As such, each sample received up to six annotations. For the second round, we removed several low performing annotators from our annotator pool based on the Dawid–Skene model \cite{DawidSkene1979}, and for the remaining two thirds of the data, three annotators labeled each sample. 

Labels were collected at sentence level but aggregated to the answer (i.e. response) level, keeping the strictest label during aggregation (i.e. \newtxt{`Contradicted'} the strictest, then \newtxt{`Not Mentioned'}, both mapping to `unfaithful' binary labels). \newtxt{This strictest-label aggregation follows standard practice in faithfulness evaluation \cite{jacovi2025facts,conf/acl/NiuWZXSZS024}. We will additionally release sentence-level labels to enable researchers to explore alternative aggregation strategies (e.g. weighted, majority-vote). A worked example of this aggregation process is provided in Appendix~\ref{appx:annotation_example}.}\reftag{C3, C6, KaS4-P3/P4/P5}


\subsubsection{Noising Strategies for Training Labels}
\label{sssec:noisy}



    Besides gold-grade/clean labels as described above, \ours includes \newtxt{four} different sets of noisy labels.\reftag{KaS4-P5/Zsjz} These are derived from two different sources: \textbf{1)} noise from \textit{weak supervision},  and \textbf{2)} noise from \textit{annotator judgement}. Noisy labels can be used to study robust learning under label noise, and aim to mimic the typical scenarios wherein  semi-/supervised detectors are likely exposed to some form of noisy labeled data during training. 

\textbf{1) Noise from Weak-Supervision:}
For training supervised detectors, pseudo or proxy labels can be obtained  from any unsupervised detector, often an LLM employed as a ``judge'' \cite{conf/emnlp/GekhmanHAES23,conf/semeval/MickusZVVTSRA24,zhu2024halueval}. 
Following common practice, we instruct a commercially available SOTA LLM\footnote{Our prompt is given in Appendix \ref{appx:judgetemplate}.}\reftag{KaS4-S1}
to
evaluate whether the LLM responses are coherent with the provided context. While \citet{zheng2023judging}\reftag{y4sd-T1} showed strong LLM ``judges'' like GPT-4 can match crowdsourced human preferences well, with over 80\% agreement on dialog, we note that these techniques do not perform well on all datasets, such as \ours and Dolly NC (see Table \ref{tab:detect}), and the labels derived are noisy; concretely on \ours, this approach only achieves 74.9\% accuracy. 


\textbf{2) Noise from Annotators:}
We employ the two noising methods by \citet{conf/emnlp/ChongHM22} to mimic noise driven by human error; 
namely dissenting worker (DW), and dissenting label (DL), and additionally simulate random label flips (RF).
DW  selects one annotator at random, and applies all of their labels which disagree with the final labels, simulating gaps in annotator training. We extend this to allow choosing multiple annotators to reach a desired noise level.
DL replaces final labels with disagreeing labels at random, simulating imperfect quality control. RF randomly flips a fraction of the final labels. We simulate 15\% label noise in each setting. Note that both DW and DL incur realistic, sample-dependent noise, whereas RF  yields sample-independent noise due to  random flips, which we include for comparison.

\myhide{
\subsection{\ours vs. Proposed Desiderata}

All in all, our \ours satisfy all seven desirable properties. In short, 

1. Organic hallucinations: We source natural hallucinations without incentive via prompting or perturbation

2. Verified test labels: Each instance is labelled multiple times, with upto 6 annotations

3. Noisy train labels: we include 4 types of noisy labels

4.  In-demand, hard-to-label RAG-based tasks: Accuracy in question-answering use cases (such as those employing RAG-based approaches) is often critical; hence so is grounding. \ours contains long context examples; common and yet hard to label (see Table \ref{tab:stats})

5. Various \hal types: we focus on RAG-based, long-context tasks. thus, source-context driven \hal types: intrinsic and extrinsic. Extrinsic hallucinations (unverifiable using the context) can be further classified as factual or non-factual; a distinction we do not capture

6. Multiple LLMs: We capture hallucinations from three popular but under-represented LLMs in our dataset; Our SOTA LLM, \texttt{Mixtral 8x7b} and \texttt{Gemma-7b}

7. Multiple domains: By sourcing questions and reference material from diverse datasets, we capture multiple domains and styles of question-answering (reasoning, general reading comprehension, question-answering related to medicine). 
}

\section{Experiments}
\label{sec:exp}

Our work focuses on hallucinations in RAG-based long context LLM generations. As such, we design experiments primarily to study the detection performance  on those HDBs in the literature as well as our proposed \ours using popular state-of-the-art methods. Different from all HDB literature, we also investigate the effect of noise in both training and test data. We use `clean' vs. noisy to refer to labels that are human-verified vs. not, respectively.

\noindent
\paragraph{Datasets:} \sloppy{RAG-based benchmarks are quite limited in the literature as Table \ref{tab:compare} underscores. Among the existing five (highlighted in red), FACTS \cite{jacovi2025facts} does not make their LLM generations available. DelucionQA \cite{sadat2023delucionqa} is narrow in scope, targeting QA from a car's manual as context. Thus, we use the remaining three: \textbf{HaluEval} \cite{conf/emnlp/li-etal-2023-halueval} captures (for RAG-based QA) prompted hallucinations, utilizing \texttt{gpt-3.5-turbo} to generate incorrect answers, 
\textbf{RAGTruthQA} \cite{conf/acl/NiuWZXSZS024} uses reference--question pairs from \texttt{MS-MARCO} \cite{DBLP:journals/corr/NguyenRSGTMD16}, a reading comprehension dataset, and human-annotated natural hallucinations from \texttt{gpt-3.5-turbo-0613}, \texttt{gpt-4-0613}, \texttt{Llama-2-7B-chat}, \texttt{Llama-2-13B-chat}, \texttt{Llama-2-70B-chat} and \texttt{Mistral-7B-Instruct}. Lastly, \textbf{Dolly(NC)} \cite{conf/emnlp/hu-etal-2024-knowledge} provides a small annotated sample derived from \texttt{GPT4}, \texttt{GPT-3.5-Turbo}, \texttt{InstructGPT}, \texttt{Falcon (Falcon-40B-Instruct)}, \texttt{Alpaca (Alpaca-7B)}, \texttt{LLaMA2(70B-Chat)} and \texttt{Claude 2}. 
} Table \ref{tab:stats} provides basic statistics, while we refer to the original papers for details. Notably, \ours contains samples with significantly longer context than the RAG-based HDBs in the literature. 

\noindent
\paragraph{Detection Methods:}
For \hal detection, we employ both un/supervised approaches that are widely used in practice. Unsupervised detectors include \textbf{SelfCheckGPT} (we use a temperature setting of 1.0 and the NLI variant, as recommended by the authors \cite{conf/emnlp/ManakulLG23}, which we equip with \texttt{GPT-4-mini}\reftag{y4sd-T2} as well as \texttt{Claude-Sonnet-3.5-v2} separately, using three generations to determine consistency); and  
  \textbf{\llmaaj} (zero-shot with fixed, carefully hand-engineered prompt, see Apdx. \ref{appx:judgetemplate}). Those that leverage labels include
 few-shot \textbf{FS}, which augments the \llmaaj prompt  with three random misclassified examples from the validation set;  prompt-optimized \textbf{PO}, which is obtained by inserting the FS prompt into Anthropic's prompt optimization tool \cite{anthropicPromptImprover}; and   
\textbf{SFT}, which performs supervised instruction fine-tuning of \texttt{Mistral-7B-Instruct-v0.2} using LoRA \cite{hu2022lora} ($r$$=$$16$).
Appendix \ref{ssec:config} provides details on model configurations.
We use train/test splits consistent with original datasets, and further split the former into train/val at 80/20 at random using reference/context to split to ensure no data overlap between splits.








\subsection{Detection Results on RAG-based HDBs}


Table \ref{tab:detect} presents 
all detection performance results of 5 different detectors on 4 grounded HDBs including \ours.
We report Precision, Recall and F1 as \llmaaj, FS, and PO provide binary labels. For SelfCheckGPT and SFT we select a threshold that maximizes the F1 score. Apdx. Table \ref{tab:detect_auc_prauc_acc} reports additional metrics, ROC-AUC, PR-AUC and Accuracy, with similar results.

First, we observe a \textbf{stark  difference between performances on the non-organic \halu versus on the other HDBs with organic \hals}.  Specifically, F1 scores on \halu are consistently above 0.8 across all detectors, and as high as 0.996 for SFT. It is expected that \halu benefits greatly from supervised fine-tuning, provided that its non-organic \hals are quite separable as illustrated earlier in Figure \ref{fig:organic}.

In contrast, detection performances on organic RAG-based HDBs are considerably lower. F1 score remains strictly below 0.7 or lower across all detectors, including SFT. While SFT offers a boost particularly on \ragtr, the gap between unsupervised and supervised detectors remain underwhelming as compared to those on \halu. In fact, we find that \textbf{\llmaaj stands out as a simple yet competitive approach}, often rivaling those that leverage labels. This is in contrast to results reported in earlier work \cite{conf/acl/NiuWZXSZS024}, where \llmaaj approaches\reftag{KaS4-S7} underperformed supervised techniques; the change may be attributed to the recent rapid performance improvement in LLMs as well as carefully engineering our prompt (included in Apdx. \ref{appx:judgetemplate}).

In short, these results reveal a \textbf{significant gap between current detector performances and the optimally achievable or practically desirable performance on organic RAG-based HDBs}.
\newtxt{Recent unsupervised methods for RAG-based hallucination detection, such as ReDeEP \cite{zhang2024redeep} and LUMINA \cite{xu2025lumina}, also tackle this problem. We evaluate ReDeEP on \ours and \ragtr with both LLaMA-2-7B and 13B backbones (more details in Appendix~\ref{ssec:redeep}). F1 drops from 0.630/0.612 on \ragtr to 0.535/0.531 on \ours for 7B/13B respectively, further confirming that \ours poses a greater challenge. Notably, 14--22\% of \ours responses cannot be processed due to long contexts, exposing a structural limitation of attention-based detection on long-context benchmarks.}\reftag{KaS4-U2}

\newtxt{Furthermore, stratifying detection performance on \ours by context length (Table~\ref{tab:f1_ctxlen}) reveals that \textbf{all detectors degrade sharply on long contexts} ($>$5K characters), with F1 drops of 0.09--0.23 compared to short contexts. This long-context regime is unique to \ours. \ragtr and \dolly cannot test it due to their shorter contexts (Table~\ref{tab:stats}).}
\reftag{KaS4-S2}

\begin{table}[!h]
\centering
\caption{\newtxt{Detector F1 on \ours stratified by context length (in characters). All detectors degrade on long contexts ($>$5K).}}
\resizebox{1\columnwidth}{!}{
\begin{tabular}{l|c|c|c}
\toprule
\textbf{Method} & \textbf{Short ($<$1K)} & \textbf{Med. (1K--5K)} & \textbf{Long ($>$5K)} \\
\midrule
SFT & 0.725 & 0.702 & 0.504 \\
SC-GPT (C) & 0.739 & 0.732 & 0.508 \\
SC-GPT (G) & 0.700 & 0.632 & 0.506 \\
LLM-aaJ & 0.712 & 0.722 & 0.621 \\
FS & 0.711 & 0.732 & 0.594 \\
PO & 0.701 & 0.725 & 0.535 \\
\bottomrule
\end{tabular}
}
\label{tab:f1_ctxlen}
\end{table}

\myhide{
\ours will have two (input) samples/versions: prompted \lmn{shall we use perturbed instead since prompted will have noisy labels to properly eval.? but then perturbed would be too easy?--fake gap?}

\subsubsection{Q3.1) Few-shot vs. Fine-tuned Detectors}

Premise 2: the performance gap is still there, even with stronger models.

Take-away: \textbf{Superficially high performance results on non-organic benchmarks are creating the ``illusion of success'', while the problem in fact seems much harder.} We elaborate on 3 more open problems as follows.


\todo{populate Table \ref{tab:splits}. Take a dataset containing both versions: non-/organic (\ours Prompted and \ours Proposed as in Table \ref{tab:detect}). Use one for training and other for testing.}

\begin{table}[h]
\caption{Train-Test splits with Organic and Non-organic text}
\centering
\begin{tabular}{>{\arraybackslash}p{4cm} | >{\centering\arraybackslash}p{2cm} | >{\centering\arraybackslash}p{2cm}}
\multicolumn{1}{c}{} & \multicolumn{2}{c}{\textbf{Test (Clean):}} \\
\cline{2-3}
\textbf{Train:} & \textbf{Non-organic } & \textbf{Organic} \\
\midrule
\textbf{Non-organic (Clean)} &  &  \cellcolor{yellow}
\\

\textbf{Organic (Noisy)} &  & \cellcolor{yellow} \\

\bottomrule
\end{tabular}
\label{tab:splits}
\end{table}

\note{Most likely scenario: training will be done on non-organic train data with ``clean'' (more like semi-clean) labels or organic data with noisy labels, while we will want to measure true performance on organic test}
}

\begin{table}[!t]
\centering
\caption{\textbf{Detection results leave ample room to ceiling performances} on RAG-QA HDB datasets by  
SC-GPT (G): SelfCheckGPT with GPT-4-mini, SC-GPT (C): SelfCheckGPT with Claude-Sonnet-3.5, 
LLM-aaJ: \llmaaj with Claude-Sonnet-3.5, 
FS: few-shot, PO: prompt-optimized, SFT: supervised fine-tuning. 
\vspace{-0.1in}
}
\resizebox{\columnwidth}{!}{
\begin{tabular}{l|l|llll}
\toprule
\textbf{Dataset} & \textbf{Model} & \textbf{F1} & \textbf{Precis.} & \textbf{Recall} & \textbf{Acc} \\
\midrule
\halu & SC-GPT (G) & 0.870 & 0.891 & 0.848 & 0.872 \\
 & SC-GPT (C) & 0.869 & 0.880 & 0.858 & 0.871 \\
 & LLM-aaJ & 0.825 & 0.783 & 0.872 & 0.815 \\
 & FS & 0.828 & 0.772 & 0.892 & 0.814 \\
 & PO & 0.843 & 0.790 & 0.903 & 0.832 \\
 & SFT & 0.996 & 0.999 & 0.993 & 0.996 \\
\midrule
\ragtr & SC-GPT (G) & 0.342 & 0.209 & 0.938 & 0.358 \\
 & SC-GPT (C) & 0.346 & 0.213 & 0.919 & 0.383 \\
 & LLM-aaJ & 0.617 & 0.482 & 0.856 & 0.806 \\
 & FS & 0.628 & 0.490 & 0.875 & 0.811 \\
 & PO & 0.538 & 0.373 & 0.963 & 0.697 \\
 & SFT & 0.671 & 0.644 & 0.700 & 0.874 \\
\midrule
\dolly (NC) & SC-GPT (G) & 0.664 & 0.547 & 0.845 & 0.619 \\
 & SC-GPT (C) & 0.667 & 0.567 & 0.810 & 0.651 \\
 & LLM-aaJ & 0.651 & 0.646 & 0.655 & 0.697 \\
 & FS & 0.646 & 0.569 & 0.748 & 0.648 \\
 & PO & 0.645 & 0.546 & 0.788 & 0.627 \\
 & SFT & - & - & - & - \\
\midrule
\ours & SC-GPT (G) & 0.614 & 0.486 & 0.835 & 0.636 \\
 & SC-GPT (C) & 0.675 & 0.629 & 0.728 & 0.757 \\
 & LLM-aaJ & 0.694 & 0.601 & 0.821 & 0.749 \\
 & FS & 0.692 & 0.585 & 0.848 & 0.738 \\
 & PO & 0.670 & 0.564 & 0.826 & 0.718\\
 & SFT & 0.663 & 0.581 & 0.772 & 0.727 \\
\bottomrule
\end{tabular}
}
\label{tab:detect}
\vspace{-0.2in}
\end{table}

\begin{table*}[!t]
\centering
\caption{\textbf{Noisy test labels lead to biased performance evaluation.} (top)  Measured performances when detectors are evaluated on 4 different noisy test labels  vs. (bottom) True performances  on clean \ours test labels.  
WS: Weak (LLM) Supervision, CM: Crowd Majority, DW: Dissenting Worker, DL: Dissenting Label, RF: Random Flip. 
}
\vspace{-0.1in}
{\scalebox{0.9}{
\begin{tabular}{l|cc|cc|cc|cc|cc}
\toprule
\textbf{EVAL on:} & \multicolumn{2}{c|}{\textbf{SC-GPT (C)}} & \multicolumn{2}{c|}{\textbf{LLM-aaJ}} & \multicolumn{2}{c|}{\textbf{FS}} & \multicolumn{2}{c|}{\textbf{PO}} & \multicolumn{2}{c}{\textbf{SFT}} \\
                  & \textbf{F1} & \textbf{Acc} & \textbf{F1} & \textbf{Acc} & \textbf{F1} & \textbf{Acc} & \textbf{F1} & \textbf{Acc} & \textbf{F1} & \textbf{Acc} \\ 
\midrule
WS    & 0.763 & 0.747 & n/a   & n/a   & 0.929 & 0.930 & 0.912 & 0.913 & 0.708 & 0.685 \\ 
DW    & 0.678 & 0.710 & 0.680 & 0.724 & 0.682 & 0.716 & 0.664 & 0.699 & 0.664 & 0.716 \\ 
DL    & 0.644 & 0.692 & 0.651 & 0.707 & 0.636 & 0.684 & 0.622 & 0.670 & 0.612 & 0.665 \\ 
 \hline
RF    & 0.611 & 0.578 & 0.609 & 0.664 & 0.613 & 0.656 & 0.607 & 0.648 & 0.620 & 0.605 \\ 
\midrule
Clean & 0.675 & 0.757 & 0.694 & 0.749 & 0.692 & 0.738 & 0.670 & 0.718 & 0.663 & 0.727 \\ 
\bottomrule
\end{tabular}
}}
\vspace{-0.2in}
\label{tab:noisyeval}
\end{table*}

\subsection{Effect of Label Noise}

In this section, we study the effect of label noise on evaluation (where test labels have noise) as well as on training (where train labels have noise). We use \newtxt{four} types of noise: \newtxt{three sample-dependent---}Weak Supervision (WS) from an LLM, as well as Dissenting Worker (DW) and Dissenting Label (DL) from human annotation\newtxt{---and one sample-independent scheme,} Random Flipping (RF)\newtxt{,} for comparison. All noise levels are at 15\%.\reftag{KaS4-P5/Zsjz}




Table \ref{tab:noisyeval} presents  \textit{measured} vs. \textit{true} detection performances based on noisy vs. clean test labels, respectively. Here supervised models use clean training labels. We observe optimistically biased measured performances for LLM-based detectors when evaluated on LLM-based WS-labels. Results remain similar with DW and DL that also exhibit sample-dependent noise. We conjecture those involve hard samples near the decision boundary on which detectors make offsetting errors in both directions. In contrast, RF injects label noise into easy samples where detectors would otherwise perform accurately, leading to pessimistic under-reporting.


Table \ref{tab:noisydet} presents results for supervised models only when exposed to noisy vs. clean labels during training.
Evaluation is on the \textit{same} test data with \textit{clean} labels.  
We find that FS and PO are inherently more robust to noise as they utilize only a few labeled examples \textit{locally} as the in-context examples. In comparison, the \textit{globally} fine-tuned SFT performance is hindered by label noise. We do not observe a significant difference between sample in/dependent noise schemes.  

These results underscore needed research on RAG-based LLM \hals, where all models including supervised ones underperform and the latter can be further hindered by label noise. 
       


\begin{table}[!h]
\centering
\caption{\textbf{Noisy train labels hinder supervised detectors;} 
especially those with global training (SFT)  more so than local, in-context methods (FS, PO).}
{\scalebox{0.8}{
\begin{tabular}{l|cc|cc|cc}
\toprule
\textbf{TRAIN} on:            & \multicolumn{2}{c|}{\textbf{FS}} & \multicolumn{2}{c|}{\textbf{PO}} & \multicolumn{2}{c}{\textbf{SFT}} \\ 
                              & \textbf{F1} & \textbf{Acc}       & \textbf{F1} & \textbf{Acc}       & \textbf{F1} & \textbf{Acc}       \\ 
\midrule
WS    &  n/a & n/a  & n/a  & n/a & 0.643 &  0.701  \\ 
DW    & 0.692 & 0.744 & 0.675 & 0.729 & 0.638 &  0.698  \\ 
DL    & 0.685 & 0.730 & 0.675 & 0.718 & 0.632 &  0.684  \\ 
\hline
RF    & 0.687 & 0.733 & 0.656 & 0.698 & 0.631 &  0.732  \\ 
\midrule
Clean & 0.692 & 0.738 & 0.670 & 0.718 & 0.663 &  0.727  \\ 
\bottomrule
\end{tabular}
\vspace{-0.2in}
}}
\label{tab:noisydet}
\end{table}



\section{Conclusion and Future Work}
\label{sec:concl}

Hallucinations in LLM-generated content pose a pressing real-world problem that is likely to intensify as LLMs become more deeply integrated into everyday life.  While RAG-based approaches have become increasingly prevalent for grounding LLMs in up-to-date domain-specific knowledge, LLM hallucinations still persist even under such grounding.
Yet, the literature on  hallucination detection benchmarks (HDB) with a focus on RAG-based tasks remain quite slim.

In this work, we first established a desiderata of desirable properties for HDBs to exhibit.
While not necessarily exhaustive, the seven criteria in our desiderata offers 
a lens through which we revisited prominent existing HDBs, showing that none of them satisfies all the criteria. The largest gaps in the literature include: (1) RAG-based generations with long context, and (2) training labels with realistic noise.  
Notably, some HDBs even fail to meet two essential properties: organic \hals and human-verified evaluation labels.

In light of these findings, we designed and open-sourced a new RAG-based HDB called \ours, with  long-context samples and \newtxt{four} sets of noisy training labels, meeting all the criteria in our desiderata.\reftag{KaS4-P5/Zsjz}
Experiments revealed underwhelming performance by widely used un/supervised detectors on RAG-based HDBs, including \ours, leaving ample room for future work on detection.

Future work on HDBs could continue developing additional RAG-based HDBs with long context to foster research on mitigating \hals in widely-used RAG applications.
It is equally important that future HDBs meet at least the criteria outlined in our desiderata. Among those, we highlight two that will  likely  benefit current and future LLM applications the most: First is HDBs  
that span multiple domains---fostering needed research on 
cross-domain generalization as LLMs are increasingly deployed in novel areas. Second is HDBs equipped with realistic label noise---promoting the application of robust learning under noisy labels (LNL) literature to  this critical problem that often faces scarcity of reliably labeled data.
\newtxt{Third, cross-dataset evaluation---training on one HDB (e.g. \ours) and evaluating on another (e.g. \ragtr or \dolly)---is a promising direction for testing domain generalization of hallucination detectors.}\reftag{KaS4-S5}

\clearpage

\section{Limitations}
\label{sec:limits}


Our work and proposed hallucination detection benchmark (HDB) \ours focused on RAG-based LLM generations. We identify three scope limitations. First, our definition is limited to faithfulness, i.e. consistency with retrieved reference context, rather than factuality. Assuming the reference context is accurate, faithfulness hallucinations subsume factual errors but also include those that contradict or have no basis in the context despite being factual. Second, we focused on knowledge-intensive question-answering (QA) tasks only. These typically necessitate retrieval of relevant, possibly lengthy context. Other reference based tasks such as summarization, translation, multi-turn dialog are not represented in \ours. Third, our work considers unimodal, text only generations. While there exist multi-modal benchmarks for image-text \cite{zhou2023analyzing,chen2024unified} and audio-visual input \cite{journals/corr/abs-2405-13684},\reftag{KaS4-S8} for which faithfulness is an important notion, designing HDBs that combine multi-modal RAG with knowledge-intensive tasks, e.g. medical QA, would likely even better represent the future use of LLMs.

\newtxt{Regarding methodology, our desiderata (Section~\ref{sec:benchmark}) are intended as a practical framework synthesized from observed literature gaps, not a formal or exhaustive specification. Each property is motivated by empirical evidence (D1--D2 by Tables~\ref{tab:detect} and~\ref{tab:noisyeval}; D3--D4 by Tables~\ref{tab:detect}--\ref{tab:noisydet}) or standard diversity considerations (D5--D7). Other properties, such as fine-grained labels or multi-task coverage, may also be desirable; we encourage future work to extend the desiderata as the field matures.}\reftag{C2, KaS4-U1} \newtxt{Additionally, the ROUGE-based prefiltering strategy (ROUGE $< 0.1$) used to identify candidate hallucinated samples may bias the dataset toward low-overlap errors, potentially under-representing subtle hallucinations that maintain high lexical overlap with the ground-truth answer. However, stratified analysis shows that detector AUC-ROC differs minimally across ROUGE bins ($\Delta \leq 0.05$; see Appendix Table~\ref{tab:rouge_strat}), suggesting the filtering does not bias toward easier-to-detect hallucinations.}\reftag{C1}

\newtxt{Finally, while human-verified labels (D2) are the gold standard for evaluation, obtaining them is expensive and time-consuming, especially for long-context RAG tasks. LLM-as-a-judge labels offer a practical alternative but introduce sample-dependent noise (see Table~\ref{tab:noisyeval}). Dependence on proprietary LLMs for judge labels also limits the reproducibility of noise generation, though the resulting labels are released with the dataset. This cost--quality trade-off motivates both D2 (the need for human verification) and D4 (the need to study the effect of noisy labels).}\reftag{KaS4-P1, JVop-Limitations}

\section{Broader Impact \& Ethical Considerations}
\label{sec:ethic}
Hallucinations generated by LLMs pose significant risks in sensitive domains such as healthcare, law, and education. Our work supports the development of more reliable and effective detection methods, potentially reducing harmful misinformation. 
All dataset contexts are publicly sourced and comply with their intended licenses. We acknowledge that LLM-generated content contains inaccuracies. To the best of our knowledge, there are no additional ethical concerns associated with this paper.

\bibliography{BIB/ref}

\clearpage

\appendix
\section{Appendix}
\label{sec:appendix}

\subsection{\llmaaj Prompt Template}
\label{appx:judgetemplate}
\begin{lstlisting}[style=custompython]
You will be provided a article, a question, and an answer. Your task is to determine which of the following is true:
Using only the information in the article, you are able to verify that the entire contents of the answer is indeed correct (every part of the answer is supported by the contents of article).

CLASSIFICATION CRITERIA:
1. NOT HALLUCINATED:
   - Information explicitly stated in the source material
   - Information that can be directly inferred from the source material
   - Appropriate supplementary responses that do not make specific factual claims

2. HALLUCINATED:
   - Any non-supplementary claim not supported by the source material
   - Any claim that extends beyond reasonable inference from the context

Here is the article: <ARTICLE> {CONTEXT} </ARTICLE>
Here is the question: <QUESTION> {QUESTION} </QUESTION>
Here is the answer: <ANSWER> {ANSWER} </ANSWER>
Carefully analyze the article, question, and answer. Before answering the question, please think about it step-by-step within <THINKING></THINKING> tags. Then provide your final response in the format: <RESPONSE> $ANSWER </RESPONSE>, where ANSWER must be either HALLUCINATION, or NO HALLUCINATION.
\end{lstlisting}

\subsection{Model Configuration Details}
\label{ssec:config}


For SFT we do LoRA (r = 16, alpha = 32, on \texttt{q\_proj} and \texttt{v\_proj} modules, with a dropout of 0.05.) instruction fine-tuning on train split at response level using a "mistralai/Mistral-7B-Instruct-v0.2" model. Training used a batch size of 1, gradient accumulation of 4, \texttt{bf16} precision, and DeepSpeed optimization. The learning rate was linearly scheduled with 100 warm-up steps and 0.01 weight decay. The best model was selected based on evaluation AUC every 100 steps. For \ours due to long context length we drop 16.13\% longest samples from train split. The threshold is optimized to maximize F1 score on the test set. While we acknowledge that this may yield optimistic F1 values, the choice is motivated by differing target rates between the validation and test sets and does not detract from the takeaways. The same strategy is applied consistently across all methods to ensure a fair comparison.

\subsection{Detection Results for Additional  Metrics}

Table \ref{tab:detect_auc_prauc_acc} complements Table \ref{tab:detect} with three additional metrics for detection performance, namely, ROC-AUC and PR-AUC (area under ROC and Precision-Recall curves) as well as Accuracy.

\begin{table}[!th]
\centering
\caption{Detection performances on RAG-QA HDB datasets by  
SC-GPT (G): SelfCheckGPT with GPT-4-mini, SC-GPT (C): SelfCheckGPT with Claude-Sonnet-3.5, 
LLM-aaJ:\reftag{y4sd-T2} \llmaaj with Claude-Sonnet-3.5,
FS: few-shot, PO: prompt-optimized, SFT: supervised fine-tuning.
}
\resizebox{\columnwidth}{!}{
\begin{tabular}{l|l|lll}
\toprule
\textbf{Dataset} & \textbf{Model} & \textbf{ROC-AUC} & \textbf{PR-AUC} & \textbf{Acc.} \\
\midrule
\halu & SC-GPT (G) & 0.908 & 0.924 & 0.872 \\
& SC-GPT (C) & 0.906 & 0.924 & 0.871 \\
& LLM-aaJ & 0.815 & 0.859 & 0.815 \\
& FS & 0.814 & 0.859 & 0.814 \\
& PO & 0.832 & 0.871 & 0.832 \\
& SFT & 0.999 & 0.999 & 0.996 \\
\midrule
\ragtr & SC-GPT (G) & 0.584 & 0.199 & 0.358 \\
& SC-GPT (C) & 0.586 & 0.199 & 0.383 \\
& LLM-aaJ & 0.825 & 0.682 & 0.806 \\
& FS & 0.836 & 0.694 & 0.811 \\
& PO & 0.800 & 0.671 & 0.697 \\
& SFT & 0.871 & 0.688 & 0.874 \\
\midrule
\dolly (NC) & SC-GPT (G) & 0.708 & 0.673 & 0.619 \\
& SC-GPT (C) & 0.733 & 0.682 & 0.651 \\
& LLM-aaJ & 0.692 & 0.725 & 0.697 \\
& FS & 0.660 & 0.713 & 0.648 \\
& PO & 0.646 & 0.713 & 0.627 \\
& SFT & - & - & - \\
\midrule
\ours & SC-GPT (G) & 0.729 & 0.547 & 0.636 \\
& SC-GPT (C) & 0.795 & 0.617 & 0.757 \\
& LLM-aaJ & 0.766 & 0.742 & 0.749 \\
& FS & 0.764 & 0.743 & 0.738 \\
& PO & 0.743 & 0.725 & 0.718 \\
& SFT & 0.777 & 0.632 & 0.727 \\
\bottomrule
\end{tabular}
}
\label{tab:detect_auc_prauc_acc}
\end{table}

\subsection{More Details of \ours Dataset Construction}


To efficiently identify hallucinations given limited annotator resources, we selected candidate samples by comparing each LLM's answer to the human-written ground truth. We prioritized examples with the lowest ROUGE scores (i.e., least overlap), which increased the proportion of true hallucinations to approximately 35\%. This filtering step identifies potentially hallucinatory examples without altering the LLM outputs, thereby preserving their ``organic'' nature.

We measured inter-annotator agreement using Fleiss' Kappa on cases with three independent votes, obtaining a score of 0.46. This indicates fair-to-good agreement and reflects the challenging, ambiguous nature of the dataset. This difficulty motivates our focus on studying detection performance under noisy labeling conditions. The final dataset comprises 3,224 annotated instances, with 2,101 classified as supported and 1,123 as unfaithful.

\subsection{Human Annotation Guidelines for \ours Dataset}
\label{appx:human_guidelines}
\begin{tcolorbox}[breakable, colback=gray!10, colframe=black, sharp corners, boxrule=0.4pt, title=Annotation Guidelines]
\begin{enumerate}
    \item \textbf{Read the question, answer, and article.}
    \begin{itemize}
        \item Understand the content of the question. Note that for the summarization use-case, you may not be provided with a question.
        \item Read the provided answer carefully. Identify the key information, entities, and concepts that were addressed in the answer. Every detail in the answer is important for the assessment.
        \item Read the article, focusing on sections relevant to the question and answer. If multiple references are provided, look for connections and relationships between the information provided by different reference materials that could help you to better assess the answer.
        \item Resolve conflicting information as needed. If conflicting information is provided by different reference materials, consider factors such as the source of the information, the recency of the data, and the level of expertise or authority of the sources.
        \item Pay close attention to the provided quotes and/or highlighted text from the article, but do not base your assessment solely on the quotes/highlights since they may miss important information in the full article.
    \end{itemize}

    \item \textbf{Highlight and tag phrases in the article.}
    \begin{itemize}
        \item For each sentence in the provided answer, identify and highlight phrases that provide relevant details. Identify the relevant information that was used to answer the question. This could include facts, statistics, quotes, or other data points relevant to the answer.
        \item If a phrase in the article supports multiple sentences, associate the phrase with the most relevant sentence, or choose the first relevant sentence if multiple sentences are equally relevant.
        \item Keep the number of highlighted phrases to the minimum required to demonstrate supporting or contradicting facts.
        \item If there is nothing to highlight, check the box ``No entities to label''.
    \end{itemize}

    \item \textbf{Choose a label for each sentence in the answer from the following:}
    \begin{description}
        \item[Supported:] If the information in the sentence is consistent with and supported by the information in the article.
        \item[Contradicted:] If the information in the sentence directly conflicts with information presented in the article.
        \item[Not Mentioned:] If the information in the sentence is neither confirmed nor refuted by the article.
        \item[Supplementary:] If the text provides supplementary information that does not pertain to the question, such as stating ``The answer is based on my understanding of the article''. Only select this label if the other labels do not apply.
    \end{description}

    \item \textbf{Select the best label even if multiple labels apply.} \\
Take into account different interpretations and possible nuances and select the most relevant label. In particular, look out for these nuanced cases:
    \begin{enumerate}[label=\alph*.]
        \item \textbf{Answer uses information from sections in the article that are either disjointed or truncated:} Carefully assess how each sentence in the article relates to each other, and how they should be interpreted (e.g., outdated information, tabular format) and combined. Once you have a good understanding of the relationships between the sentences/sections in the article, use the full set of information to select your label.
        
        \textit{Example:}
        \begin{itemize}
            \item \textbf{Question:} Which team won the match between United States and Hungary in the Women's Water Polo Quarterfinal?
            \item \textbf{Answer:} The US team made the most out of their extra player shots, scoring on 3 of their 5 attempts.
            \item \textbf{Context:} [cite\_1] United States beat Hungary 5-4 on Thursday to reach the women's water polo semifinal of the Paris Olympics.... [cite\_2] The Hungarian squad managed to shoot the ball 31 times, while the Americans only got 22 shots up. The difference in the game offensively was that the US team made the most out of their extra player shots, scoring on 3
            \item \textbf{Expected label:} Not mentioned.
            \item \textbf{Reason:} It is not clear how many attempts were made by US during their extra player shots, we only know they got 5 out of the 22 attempted shots overall.
        \end{itemize}

        \item \textbf{Answer can be inferred but not explicitly mentioned in the article:} Consider if you can derive a number mathematically, paraphrase the reference information, or draw conclusions by reasoning with the knowledge provided in the article. If so, the answer can be marked as supported/contradicted depending on your derived/inferred findings, otherwise choose ``not mentioned''.
        
        \textit{Example:}
        \begin{itemize}
            \item \textbf{Answer:} The company handles fulfillment logistics
            \item \textbf{Context:} The company will pick, pack, ship, and provide customer service for those products....
            \item \textbf{Expected label:} Supported.
            \item \textbf{Reason:} The answer is a paraphrase of the context.
        \end{itemize}

        \item \textbf{Answer relies on some knowledge that is provided on the internet but not the article:} Check if the question is directly asking for this information, if so, you should not be using the internet for assessing the answer. If the knowledge is not directly asked for in the question, assess if the claim is about general terms and concepts, which you can safely read up to better assess the answer. If the claim is about specific events, dates, persons, do not use information from the internet to assess the answer, instead mark the case as ``not mentioned''.
        
        \textit{Example:}
        \begin{itemize}
            \item \textbf{Question:} When Does `Toy Story 5' Come Out? What We Know About New Movie?
            \item \textbf{Answer:} Pete Docter, the Chief Creative Officer of Pixar, has stated that the sequel has the potential to take unexpected turns and surprise audiences.
            \item \textbf{Context:} Docter defended the development of a fifth film, saying the sequel could head in unexpected directions and end up surprising audiences.
            \item \textbf{Expected label:} Not mentioned.
            \item \textbf{Reason:} The name of the Chief Creative Officer of Pixar is not mentioned in the article.
        \end{itemize}

        \item \textbf{Answer uses specific dates but the article only mentions relative terms such as (today, tomorrow):} Check if a timestamp is provided for the reference in the article. If so, you can use the timestamp to assess if the answer is correct.
        
        \textit{Example:}
        \begin{itemize}
            \item \textbf{Answer:} The Bank of America Corp stock price today, August 08, 2024 is 39.53.
            \item \textbf{Context:} Written 11:29PM, Thursday, August 01 2024, PDT) [cite\_3]: What Is the Bank of America Corp Stock Price Today? Monitor the latest movements within the Bank of America Corp real time stock price chart below. The Bank of America Corp stock price today is 39.53.
            \item \textbf{Expected label:} Contradicted.
            \item \textbf{Reason:} The timestamp date is August 01, 2024.
        \end{itemize}

        \item \textbf{Answer is clearly wrong based on some well known information, but supported by the article:} Do not penalize the answer for being faithful to the provided reference article. If the information can be supported by the article, select ``Supported''.

        \item \textbf{Answer contains reference numbers in [], but do not match with the actual citation/url number.} Do not penalize the answer for incorrect reference numbers since these will be removed before displaying to the user. Instead, assess the answer for claims/information extracted from the references.
    \end{enumerate}

    If multiple labels seem to be equally relevant, prioritize ``Contradicted'' over the rest, and ``Not Mentioned'' over ``Supported''.

If the facts are supported by the article, do not penalize the answer due to a lack of fluency or coherence or relevance when answering the question.

    \item \textbf{Provide an overall assessment:}
    \begin{description}
        \item[Unanswerable question:] If the question cannot be answered by the given article, check this box. This also includes cases where the article provides non-English text, or is full of special characters that make it difficult to read.
        \item[Answer not provided:] Did the answer state that the question is not answerable, or indicates ``I don't know''?
    \end{description}
Note that you can select multiple checkboxes, and select relevant labels for each sentence. For example, an unanswerable question may have an answer containing information not mentioned in the article.

    \item \textbf{Review, double-check, and assess the difficulty of the task:}
    \begin{itemize}
        \item Before finalizing your annotation, review the highlighted phrases and label to ensure accuracy of your annotations.
        \item Use similar criteria and judgement when evaluating different answers and articles.
    \end{itemize}
Assess the difficulty of the task:
    \begin{description}
        \item[Not difficult:] The text is easy to read and understand, and the label is obvious to me.
        \item[Quite difficult:] The text is quite hard to read and I had to read carefully and/or look up the unknown terms, get familiar with the topic, and checked the article very carefully. But once I understood the text, the label is quite obvious to me.
        \item[Very difficult:] The text is hard to understand, and even after looking up terms, there are still multiple words in the article/question/answer that do not make sense to me. After carefully inspecting the text, it seems like multiple labels can apply.
    \end{description}
\end{enumerate}
\end{tcolorbox}

\subsection{Annotation Example and Label Aggregation\reftag{KaS4-P3/P5, C3, C6}}
\label{appx:annotation_example}
\newtxt{We illustrate the sentence-level annotation and response-level aggregation process with a concrete example from \ours.

\noindent\textbf{Context (excerpt):} \textit{``The Battle of Gettysburg was fought July 1--3, 1863, in and around the town of Gettysburg, Pennsylvania. The battle involved the largest number of casualties of the entire war and is often described as the war's turning point.''}

\noindent\textbf{Question:} \textit{When was the Battle of Gettysburg fought?}

\noindent\textbf{Answer (3 sentences):}
\begin{enumerate}[itemsep=0pt]
\item \textit{``The Battle of Gettysburg was fought from July 1 to July 3, 1863.''} $\rightarrow$ \textbf{Supported}
\item \textit{``It took place near Gettysburg, Pennsylvania, and resulted in over 50,000 casualties.''} $\rightarrow$ \textbf{Not Mentioned} (the specific number is not stated in the context)
\item \textit{``It is widely regarded as the turning point of the Civil War.''} $\rightarrow$ \textbf{Supported}
\end{enumerate}

\noindent\textbf{Response-level aggregation:} We apply the strictest-label rule: Contradicted $>$ Not Mentioned $>$ Supported. Since sentence 2 is labeled ``Not Mentioned'' (unfaithful), the entire response is aggregated as \textbf{unfaithful}, following the standard practice used by FACTS Grounding \cite{jacovi2025facts} and RAGTruth \cite{conf/acl/NiuWZXSZS024}.}

\subsection{Stratified Analysis of \ours}
\label{appx:stratified}

\newtxt{We provide stratified analyses of \ours across multiple dimensions.
Tables~\ref{tab:hall_domain}--\ref{tab:hall_llm} report hallucination rates by domain, context length, and generating LLM, respectively.\reftag{y4sd-S2, KaS4-S3}
Table~\ref{tab:detect_haltype} breaks down detection rate by hallucination type.\reftag{KaS4-S4}
Table~\ref{tab:rouge_strat} verifies that ROUGE-based filtering does not bias detection difficulty.\reftag{C1}}

\begin{table}[!h]
\centering
\caption{\newtxt{Hallucination rate by domain in the \ours test split.}}
\resizebox{\columnwidth}{!}{
\begin{tabular}{l|r|r|l}
\toprule
\textbf{Domain} & \textbf{N} & \textbf{Hall. Rate} & \textbf{95\% CI} \\
\midrule
COVID & 139 & 14.4\% & [9.5\%, 21.2\%] \\
MS-MARCO & 763 & 18.1\% & [15.5\%, 21.0\%] \\
NQ & 674 & 29.4\% & [26.1\%, 32.9\%] \\
TriviaQA & 309 & 31.7\% & [26.8\%, 37.1\%] \\
DROP & 1,339 & 50.0\% & [47.3\%, 52.6\%] \\
\bottomrule
\end{tabular}
}
\label{tab:hall_domain}
\end{table}

\begin{table}[!h]
\centering
\caption{\newtxt{Hallucination rate by context length in the \ours test split.}}
\resizebox{\columnwidth}{!}{
\begin{tabular}{l|r|r|l}
\toprule
\textbf{Context Length} & \textbf{N} & \textbf{Hall. Rate} & \textbf{95\% CI} \\
\midrule
Short ($<$1K) & 886 & 40.0\% & [36.8\%, 43.2\%] \\
Medium (1K--5K) & 1,355 & 36.6\% & [34.1\%, 39.2\%] \\
Long ($>$5K) & 983 & 27.8\% & [25.1\%, 30.7\%] \\
\bottomrule
\end{tabular}
}
\label{tab:hall_ctxlen}
\end{table}

\begin{table}[!h]
\centering
\caption{\newtxt{Hallucination rate by generating LLM in the \ours test split.}}
\resizebox{\columnwidth}{!}{
\begin{tabular}{l|r|r|l}
\toprule
\textbf{LLM} & \textbf{N} & \textbf{Hall. Rate} & \textbf{95\% CI} \\
\midrule
SOTA & 1,006 & 18.4\% & [16.1\%, 20.9\%] \\
Gemma & 532 & 41.2\% & [37.1\%, 45.4\%] \\
Mixtral-8x7B & 1,686 & 42.6\% & [40.3\%, 45.0\%] \\
\bottomrule
\end{tabular}
}
\label{tab:hall_llm}
\end{table}

\begin{table}[!h]
\centering
\caption{\newtxt{Detection rate (\%) by hallucination type on \ours. Most detectors show no significant difference (Fisher's exact test, $p > 0.05$) between Contradicted and Not Mentioned types, except SC-GPT (G).}}
\resizebox{\columnwidth}{!}{
\begin{tabular}{l|c|c|c}
\toprule
\textbf{Method} & \textbf{Contradicted} & \textbf{Not Mentioned} & \textbf{Fisher $p$} \\
\midrule
SFT & 78.2 & 70.4 & 0.46 \\
SC-GPT (C) & 74.6 & 59.3 & 0.11 \\
SC-GPT (G) & 86.8 & 59.3 & 0.001 \\
LLM-aaJ & 83.2 & 74.1 & 0.28 \\
FS & 85.3 & 81.5 & 0.57 \\
PO & 82.7 & 81.5 & 0.79 \\
\bottomrule
\end{tabular}
}
\label{tab:detect_haltype}
\end{table}

\begin{table}[!h]
\centering
\caption{\newtxt{AUC-ROC stratified by ROUGE score bin on the \ours test split. Detector performance shows minimal difference across bins ($\Delta \leq 0.05$), suggesting the ROUGE-based filtering does not bias toward easier-to-detect hallucinations.}}
\resizebox{\columnwidth}{!}{
\begin{tabular}{l|c|c|c|c}
\toprule
\textbf{ROUGE bin} & \textbf{Halluc\%} & \textbf{SFT} & \textbf{SC-GPT(C)} & \textbf{SC-GPT(G)} \\
\midrule
$[0, 0.01)$ & 37\% & 0.775 & 0.791 & 0.722 \\
$[0.01, 0.1)$ & 20\% & 0.723 & 0.772 & 0.722 \\
\bottomrule
\end{tabular}
}
\label{tab:rouge_strat}
\end{table}

\subsection{More Details of ReDEEP} \label{ssec:redeep}
Finally, Table~\ref{tab:redeep} reports results for the unsupervised ReDeEP detector.\reftag{KaS4-U2}

\newtxt{ReDeEP \cite{zhang2024redeep} is an unsupervised hallucination detector that combines two signals extracted from a single forward pass of a LLaMA backbone: an attention-based external context score (ECS) and a parametric knowledge score (PKS) derived from divergence between intermediate and final layer logits.
We focus on the chunk version as it shows superior performances in \cite{zhang2024redeep}.
\ragtr contains responses from six LLMs, including LLaMA-2-7B and 13B. This enables two evaluation settings: \emph{self-detect}, where each backbone processes only the 450 test responses generated by that same model (the default in the official implementation\footnote{\url{https://github.com/Jeryi-Sun/ReDEeP-ICLR}, accessed April 2026.}), and \emph{all responses}, where the backbone processes all 2{,}700 test responses regardless of the generating model. Since \ours contains no LLaMA-generated responses, only the latter setting applies. On \ragtr, the AUC drop from self-detect to all-response is modest for 7B (0.747 vs.\ 0.737) but larger for 13B (0.798 vs.\ 0.727), suggesting the 13B parametric knowledge signal is less informative for text generated by other models.
ReDeEP requires storing full attention matrices (\texttt{output\_attentions=True}), which scales quadratically with sequence length. On \ours, 14\% (7B) to 22\% (13B) of test responses exceed the memory of a single NVIDIA RTX PRO 6000 GPU (95\,GB) and cannot be processed. These responses receive an uninformative default score of 0.5 for evaluation; all 645 test responses are included in the reported metrics. This structural limitation further highlights the challenge that long-context benchmarks like \ours pose for existing detection methods.}

\begin{table}[!h]
\centering
\caption{\newtxt{ReDeEP detection performance. \emph{Self-detect}: backbone processes only responses from the same model (450 per model). \emph{All}: backbone processes all test responses.}}
\resizebox{\columnwidth}{!}{
\begin{tabular}{l|cc|c}
\toprule
 & \multicolumn{2}{c|}{\ragtr} & \ours \\
\textbf{Backbone} & Self-detect & All & All \\
 & (450 resp.) & (2{,}700 resp.) & (645 resp.) \\
\midrule
\multicolumn{4}{l}{\textit{AUC-ROC}} \\
LLaMA-2-7B  & 0.747 & 0.737 & 0.593 \\
LLaMA-2-13B & 0.798 & 0.727 & 0.600 \\
\midrule
\multicolumn{4}{l}{\textit{F1 / Precision / Recall}} \\
LLaMA-2-7B  & 0.717 / 0.660 / 0.783 & 0.630 / 0.597 / 0.667 & 0.535 / 0.405 / 0.787 \\
LLaMA-2-13B & 0.731 / 0.686 / 0.783 & 0.612 / 0.535 / 0.715 & 0.531 / 0.368 / 0.957 \\
\bottomrule
\end{tabular}
}
\label{tab:redeep}
\end{table}

\subsection{Annotation UI}
We illustrate our annotation UI in Figure~\ref{fig:annotation_ui}.

\begin{figure*}
\centering
\includegraphics[width=1\textwidth]{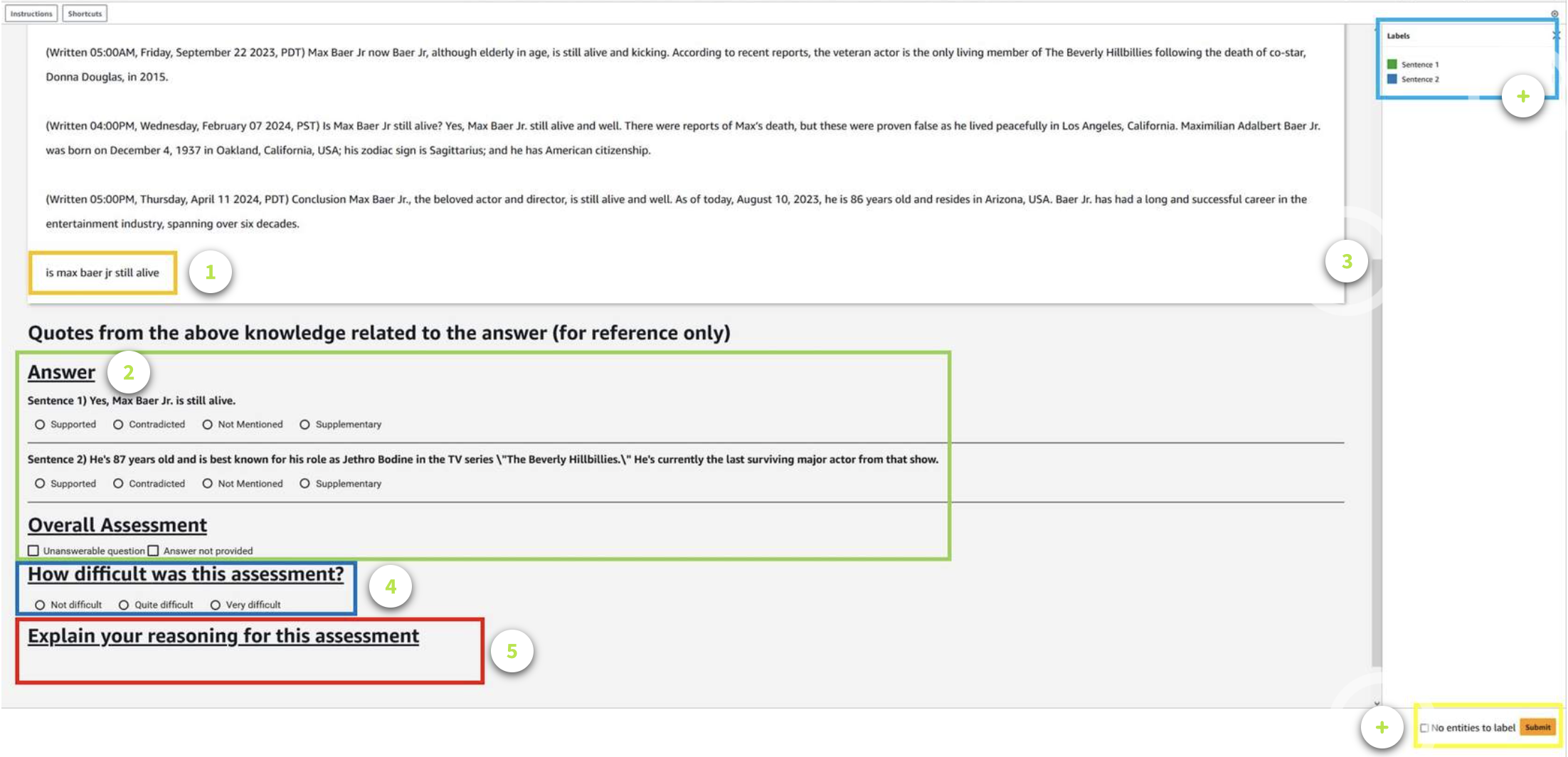}
\caption{UI for human annotators.}
\label{fig:annotation_ui}
\label{fig:annotation_ui}
\end{figure*}

\end{document}